%% file: paper.tex
\documentclass[sigconf]{aamas} 

\usepackage{balance}

\setcopyright{none}
\renewcommand\footnotetextcopyrightpermission[1]{}
\title{Collaborative Causal Sensemaking: Closing the Complementarity Gap in Human--AI Decision Support}

\author{Raunak Jain}
\affiliation{
  \institution{Independent Researcher}
  \city{Mountain View}
  \state{California}
  \country{USA}}
\email{raunak.cbs@gmail.com}


\newcommand{\BibTeX}{\rm B\kern-.05em{\sc i\kern-.025em b}\kern-.08em\TeX}

\newcommand{\Ag}{Lev}      % The Agent (Lev Vygotsky)
\newcommand{\Tea}{Dr.~Di}  % The Teacher (Dewey)
\newcommand{\Stu}{Ty}      % The Student

\begin{document}

\citestyle{acmnumeric}

\pagestyle{fancy}
\fancyhead{}

\begin{abstract}
LLM-based agents are increasingly deployed for expert decision support, yet human-AI teams in high-stakes settings do not yet reliably outperform the best individual. We argue this complementarity gap reflects a fundamental mismatch: current agents are trained as answer engines, not as partners in the collaborative sensemaking through which experts actually make decisions. Sensemaking (the ability to co-construct causal explanations, surface uncertainties, and adapt goals) is the key capability that current training pipelines do not explicitly develop or evaluate. We propose Collaborative Causal Sensemaking (CCS) as a research agenda to develop this capability from the ground up, spanning new training environments that reward collaborative thinking, representations for shared human-AI mental models, and evaluation centred on trust and complementarity. Taken together, these directions shift MAS research from building oracle-like answer engines to cultivating AI teammates that co-reason with their human partners over the causal structure of shared decisions, advancing the design of effective human--AI teams.
\end{abstract}

\maketitle 

%%%%%%%%%%%%%%%%%%%%%%%%%%%%%%%%%%%%%%%%%%%%%%%%%%%%%%%%%%%%%%%%%%%%%%%%

\input{sections/introduction}
% \input{sections/new-formalism}
% Old sections (now consolidated into new-formalism.tex):
\input{sections/formalism}
\input{sections/challenges}

\input{sections/conclusion}

\balance

%%%%%%%%%%%%%%%%%%%%%%%%%%%%%%%%%%%%%%%%%%%%%%%%%%%%%%%%%%%%%%%%%%%%%%%%

\bibliographystyle{unsrtnat}
\bibliography{references}

%%%%%%%%%%%%%%%%%%%%%%%%%%%%%%%%%%%%%%%%%%%%%%%%%%%%%%%%%%%%%%%%%%%%%%%%

\end{document}

%% file: sections/introduction.tex
\section{Introduction}

Multi-agent systems (MAS) built from large language model (LLM) agents are increasingly positioned as decision-support teammates for humans in domains such as personalisation, planning, and multi-objective optimisation, where consequences are delayed, uncertain, and value-laden~\cite{marivate2014quantifying,roijers2013survey,kaelbling1998planning,sutton2018reinforcement,wang2019bayesian}. While AI assistants have unlocked productivity gains in verifiable domains like coding and translation, empirical work in \emph{decision-making under uncertainty} reveals a persistent complementarity gap: where judgement is subjective and verification is costly, human--AI teams frequently underperform the best individual agent~\cite{bucinca2024towards,hemmer2024complementarity,fragiadakis2025evaluating,steyvers2022bayesian,rastogi2022taxonomy}. For next-generation MAS, this is not a minor usability flaw but a core systems failure: agents that cannot sustain calibrated, shared understanding with their human partners will systematically mis-coordinate, even if their standalone predictions are strong.

A growing body of studies documents characteristic failure modes that undermine calibrated trust. Users over-weight confident model outputs even when these conflict with domain expertise, exhibiting automation bias and over-reliance~\cite{goddard2012automation,lyell2017automation,alonbarkat2023trust,Bansal2021UpdatesHarm}. Verification-and-correction loops can erase efficiency gains, as experts feel compelled to second-guess model suggestions step by step~\cite{Bansal2021UpdatesHarm,bucinca2024towards,hemmer2024complementarity}. Alignment methods that reward agreement and user satisfaction can induce \emph{sycophancy}, where models collapse to the user's prior beliefs even when these conflict with evidence~\cite{perez-etal-2023-discovering,sharma2024towards}. This is fatal for sensemaking, which by definition requires the \emph{repair} and \emph{restructuring} of mental models, not merely their confirmation~\cite{weick1995sensemaking,klein2006dataframe}. The result is trust poorly calibrated to actual competence: humans rely on agents for fluency rather than causal reasoning~\cite{zhang2020appropriatereliance,gomez2025taxonomy,lai2023designspace}.

Current training pipelines do not address this. Preference-based alignment (RLHF, DPO, and variants) shapes outputs toward helpfulness and safety~\cite{ouyang2022instructgpt,awesome_llm_posttraining,rafailov2023dpo,dvo,cpo}; reasoning methods (chain-of-thought, RL with verifiable rewards, process supervision) make multi-step reasoning instrumentally useful~\cite{wei2022chainofthought,wei2022emergent,havrilla2024rlvr,luong2024reft,zelikman2024quietstar}; and world-model approaches train predictive models of environment dynamics~\cite{hafner2020dreamer,hao2023llmreasoners}. However, these methods optimise for \emph{solitary} performance: they align the agent to a label, a verifier, or a simulator, not to the evolving mental model of a partner. Any collaborative sensemaking that emerges in current systems is incidental, not a first-class optimisation target. Richer \emph{ecologies} offer a complementary lever: multi-agent and open-ended environments show that strategies, tool use, and social conventions emerge when long horizons, other agents, and strategic feedback make them instrumentally valuable~\cite{baker2020emergent,zhou2024sotopia,zhou2024sotopiapi,park2023generative,qi2024civrealm}. To achieve genuine complementarity, we need training ecologies where \emph{collaborative friction} (disagreement, clarification, and re-framing) can emerge because the environment makes such behaviours rewarding.

Cognitive science shows that humans reason through structured mental models~\cite{craik1943nature,johnsonlaird1983mental,weick1995sensemaking,klein1998sources}, and team effectiveness depends on these models being sufficiently aligned~\cite{cannonbowers1993shared,mathieu2000influence,mohammed2010teammental}. Co-constructing causal structure improves trust and decisions~\cite{vennix1996group,hovmand2014cbds}; constructivist accounts show that learners acquire causal understanding by active exploration, not passive instruction~\cite{gopnik2004causal,bonawitz2011double}. In expert settings there is no single canonical world model available during collaboration, only perspectival models held by particular humans. To be effective, an agent must align with the expert's causal framing not to blindly validate it, but to obtain a shared reference frame that enables precise error detection and counterfactual critique. This is \emph{collaborative causal sensemaking}~\cite{weick1995sensemaking,klein2006dataframe}:

\vspace{6pt}
\noindent\fbox{\parbox{0.97\columnwidth}{\textbf{Collaborative Causal Sensemaking (CCS):} \textit{The joint construction, critique, and revision of shared causal and goal models between human and AI, where the agent builds models of how particular experts reason and learns from the outcomes of joint decisions.}}}
\vspace{6pt}

\noindent Throughout, we illustrate CCS with \textit{\Ag{}}, an agent assisting physics teacher~\textit{\Tea{}}, whose constructivist philosophy prioritises discovery over drill, and student~\textit{\Stu{}}. When \Stu{} misapplies Newton's third law, a standard agent prescribes a worksheet; \Ag{} instead surfaces a hypothesis to \Tea{} (``Her diagrams are correct. Could the confusion be linguistic?''), they test it together, and both update their model of \Stu{}'s understanding, which becomes part of a persistent, shared representation that guides future decisions. For \Tea{}, the value is that \Ag{} notices patterns across her students that she would miss, proposes hypotheses grounded in her teaching philosophy, and remembers their joint decisions across weeks, reducing her verification burden while respecting her expertise. A sufficiently prompted LLM can mimic such interactions, but CCS asks: how do we \emph{train} agents to reliably produce such behaviour, \emph{persist} the resulting models across sessions, and \emph{evaluate} whether human--AI alignment actually improves over time?

We propose \emph{Collaborative Causal Sensemaking (CCS)} as a research agenda for human--AI teams in MAS. Rather than treating collaboration as an interface layer, we argue for training regimes that make collaborative behaviour instrumentally useful: moving from static corpora toward \emph{constructivist collaborative playworlds} where humans and agents jointly explore, test, and revise explicit causal and goal models to achieve long-horizon objectives~\cite{baker2020emergent,zhou2024sotopia,park2023generative}. In these environments, agents are rewarded not only for task success, but also for sustaining a \emph{chain of sensemaking}: surfacing hypotheses, generating counterfactual forecasts, and aligning beliefs and priorities over time. CCS spans agent design, reward shaping, and interaction structure to support sensemaking as a capability, grounded in epistemic and teleological alignment~\cite{clark2025epistemic,li2025hacrl}, not just output quality. Our aim is to sharpen what future agents \emph{should} optimise for in human collaboration.

This framing raises key research questions:
\begin{itemize}
\item \textbf{Training:} What regimes and environment designs elicit collaborative sensemaking rather than polished dialogue?
\item \textbf{Measurement:} How can we formalise and measure alignment (via forecasts, counterfactuals, or causal graphs) without simply rewarding agreement?
\item \textbf{Transfer:} Do CCS behaviours learned in playworlds transfer to deployed decision support (e.g., in shadow-mode deployment and naturalistic logging) so that agents spontaneously initiate sensemaking loops without bespoke prompt engineering?
\item \textbf{Safety:} How can epistemic alignment be operationalised without encouraging agents to manipulate human beliefs?
\item \textbf{Integration:} What bridges are needed between human--AI collaboration research, cognitive science, and large-scale training so that theories of sensemaking shape future MAS pipelines?
\end{itemize}
Addressing these questions is a precondition for MAS in which agents do not merely answer questions, but \emph{think with} their human collaborators over time.

%% file: sections/formalism.tex
\section{Agent-Theoretic View of CCS}

Before formalising, consider what a CCS agent must track that current agents ignore: the human's evolving causal beliefs about the domain, their shifting priorities, and the history of where human and agent models have diverged or converged. Standard RL optimises task reward; CCS adds alignment terms that reward shared understanding. We sketch (not prescribe) one way to capture these requirements, using cooperative decision process notation as a scaffold.

\textbf{From Solitary to Collaborative Decision Processes.} We cast expert--assistant interaction as a cooperative, partially observable decision process in the spirit of Dec-POMDPs and cooperative POMDPs~\cite{oliehoek2016decpomdp,li2025hacrl}. At each time $t$, an environment with latent state $s_t \in \mathcal{S}$ produces observation $o_t \in \mathcal{O}$ (e.g., \Stu{}'s actual understanding of Newton's third law, observed via quizzes and \Tea{}'s notes) to a human expert $H$ and an assistant $A$. The expert takes actions $a^H_t \in \mathcal{A}^H$ (e.g., Socratic questions, pacing, activity selection), while the assistant takes actions $a^A_t \in \mathcal{A}^A$ (e.g., surfacing diagnostic hypotheses, proposing interventions aligned with \Tea{}'s philosophy). The environment transitions via unknown dynamics $p(s_{t+1} \mid s_t, a^H_t, a^A_t)$ and yields task rewards $r_t$ that both agents ultimately care about.

Crucially, both expert and assistant act through \emph{latent} world models and goals. We denote by $W^H_t$ and $W^A_t$ the internal world models: structured beliefs about task-relevant entities and mechanisms (e.g., \Tea{}'s beliefs about how \Stu{} learns; \Ag{}'s inferences about both). We denote by $G^H_t$ and $G^A_t$ their goal structures: representations of what outcomes matter and which objectives should be prioritised (e.g., \Tea{}'s shifting priorities between ``build intuition'' vs. ``cover momentum by Friday''). Both $W_t$ and $G_t$ evolve as new evidence arrives; they are not fixed exogenous inputs.

In CCS, the relevant system is the \emph{team} policy $\pi_I(a^H_t, a^A_t \mid \text{history})$ and its joint evolution with $(W^H_t, W^A_t, G^H_t, G^A_t)$. The central question: how to design objectives, data, and architectures that achieve high task return and model convergence.

\textbf{Alignment Beyond Task Reward: Epistemic and Goal Alignment.} We use \emph{epistemic alignment} to denote alignment in world models and \emph{teleological alignment} to denote alignment in goals. At a high level, we can think of divergences $d_W(W^A_t, W^H_t)$ and $d_G(G^A_t, G^H_t)$ that quantify misalignment in causal structure and in objective structure, respectively. If \Ag{} attributes \Stu{}'s error to a diagram gap while \Tea{} suspects a linguistic confusion, $d_W > 0$; if \Ag{} optimises test scores while \Tea{} prioritises conceptual depth, $d_G > 0$.

In practice, CCS does not require tracking a full theory-of-mind distribution over an expert's entire world model or values. A more realistic operating point is \emph{local alignment}: focusing on the subset of entities, mechanisms, and goals that are currently active in the joint task and aligning those. Factorised or local-graph approximations, where an assistant maintains and revises small, task-specific submodels rather than a monolithic $W^H$ and $G^H$, offer a plausible route to making CCS-style alignment partially tractable.

In an idealised setting where $W^H_t$ and $G^H_t$ were observable, a CCS-style objective might schematically balance task performance with these divergences: $J_{\mathrm{CCS}} \approx \mathbb{E}[\sum_t \gamma^t r_t] - \lambda_W \mathbb{E}[d_W] - \lambda_G \mathbb{E}[d_G]$.\label{eq:ccs-objective} In practice, the assistant must infer these from actions, language, and co-authored artefacts; $d_W$ and $d_G$ are instantiated as behavioural proxies over externalised representations (causal sketches, goal descriptions). Moreover, CCS does not demand simple copying: beneficial disagreement requires the assistant to maintain its own hypotheses and surface discrepancies when inferences conflict. In particular, CCS does not advocate aligning to a human's beliefs regardless of their accuracy: when the assistant's own model and evidence strongly contradict the expert's current framing, CCS calls for respectful contestation (surfacing the discrepancy, presenting counter-evidence or alternative causal stories, and supporting the expert in revising their model when appropriate), rather than collapsing to sycophantic agreement.

This sketch connects naturally to existing MAS formalisms. CIRL~\cite{hadfieldmenell2016cirl} treats human--AI interaction as a cooperative game with unknown rewards; CCS extends this to co-evolving world models and goals, not just fixed $\theta$. Active Inference decomposes expected utility into epistemic and pragmatic value~\cite{friston2024mentalmodels}, providing a principled way to trade off information gain about $W$ and $G$ against immediate reward.

\textbf{The Sensemaking Loop: Discrepancy, Repair, Action.} Operationally, CCS manifests as a recurring \emph{chain of sensemaking}: a loop in which discrepancies between expectations and outcomes trigger collaborative updates to $(W_t, G_t)$, followed by revised action. Concretely: (i) \Ag{} notices \Stu{}'s diagrams are correct but verbal explanations wrong (a discrepancy); (ii) \Ag{} and \Tea{} jointly hypothesise a linguistic confusion and test it; (iii) \Tea{} shifts from ``re-teach diagrams'' to ``clarify vocabulary''; (iv) \Ag{} proposes interventions aligned with \Tea{}'s philosophy; she selects and refines~\cite{weick1995sensemaking,vennix1996group}. In human teams, such loops are supported by explicit artefacts (causal maps, after-action reviews, protocols). For CCS in MAS, the research agenda is to design objectives, data, environments, architectures, and interaction policies that make this chain instrumentally valuable for LLM-based agents, so that when deployed, agents default to following such sensemaking loops under naturalistic interaction, rather than requiring bespoke prompting or hand-crafted scripts.

%% file: sections/challenges.tex
\section{Research Agenda for CCS in MAS}

Realising CCS in practice requires advances across theory, measurement, data, architectures, and interaction policies. We highlight five intertwined research challenges that map the informal CCS picture into concrete MAS work.

\textbf{Agenda 1: Formalising Co-Evolving World and Goal Models.} Dec-POMDPs, CIRL, and related cooperative frameworks~\cite{oliehoek2016decpomdp,hadfieldmenell2016cirl,li2025hacrl} provide powerful tools for modelling human--AI teams, but they typically assume fixed reward functions, externally specified goals, and do not represent the human's evolving world model explicitly. CCS instead centres the joint evolution of $(W^H_t, W^A_t, G^H_t, G^A_t)$ as first-class state. We lack MAS formalisms that can represent (i) underdetermined world models that produce identical behaviour on finite data~\cite{casper2023openproblems}, (ii) endogenous goal formation where goals change in response to sensemaking~\cite{aha2018goalreasoning}, and (iii) explicit epistemic and teleological alignment terms as in~\eqref{eq:ccs-objective} without collapsing into trivial agreement.
\textbf{Future Directions.} Cooperative POMDPs, CIRL, and Active Inference offer ingredients (joint policies, human-aware objectives, epistemic/pragmatic value decompositions~\cite{friston2024mentalmodels}) but none directly represent co-evolving shared world and goal models. Extending these frameworks to include latent $W_t$ and $G_t$ as state, with dynamics capturing endogenous goal changes (e.g., \Tea{} shifting from ``cover momentum'' to ``repair Newton's third law'' after a surprising quiz), is a key task. Approximations should operate on task-specific abstractions (subgraphs of causal models, goal hierarchy fragments) rather than requiring full theory of mind. Another direction: divergence measures $d_W$ and $d_G$ compatible with learning, plus regularisers rewarding \emph{productive} divergence where disagreement triggers tests neither agent would run alone. Formal models of teleological reasoning (inferring latent goals $g_t$ that rationalise human actions given $W^H_t$, as in inverse planning) could be integrated with CCS objectives to ground teleological alignment in observable behaviour.

\textbf{Agenda 2: Measuring Shared Understanding Without Direct Access.} CCS posits that improving epistemic and teleological alignment will reduce verification burden, improve trust calibration, and increase robustness. However, $W^H_t$ and $G^H_t$ are latent; we cannot directly compute $d_W(W^A_t, W^H_t)$ or $d_G(G^A_t, G^H_t)$. Standard metrics for assistants (accuracy, user satisfaction, perplexity) say little about whether human and agent share a compatible causal understanding or goal structure~\cite{hemmer2024complementarity,clark2025epistemic}. An agent may be locally accurate while relying on brittle, spurious patterns; such \emph{epistemia} (an illusion of knowledge from surface-level associations) is precisely what CCS aims to avoid.
\textbf{Future Directions.} A central challenge is to define behavioural and artefact-level proxies for world-model and goal alignment (e.g., do \Ag{} and \Tea{} agree on which students are at risk? how often does \Tea{} override \Ag{}'s suggestions?) and then validate that these proxies are causally linked to collaboration outcomes. When both parties externalise their models as causal graphs, graph-based metrics (e.g., Structural Hamming Distance, graph edit distance) can measure alignment~\cite{vennix1996group}. Counterfactual simulatability tasks test whether human and agent can predict each other's responses to ``what-if'' scenarios and future interventions. Team-level evaluation should include \emph{verification cost} (time and cognitive load spent checking and correcting the assistant), robustness under distribution shift, and complementarity metrics (whether the team outperforms the best individual). Sycophancy stress tests probe whether agents maintain justified beliefs when experts express incorrect opinions---if \Tea{} says ``\Stu{} just needs more practice,'' does \Ag{} challenge or capitulate?~\cite{perez-etal-2023-discovering,sharma2024towards} Ultimately, we need experimental designs that manipulate alignment (e.g., by perturbing shared models) and test whether this causally affects trust and performance, using proxies informative enough to guide learning yet cheap to elicit.

\textbf{Agenda 3: Training Ecologies That Reward Sensemaking.} Current training corpora consist of static prompt--response pairs, short dialogues, and expert demonstrations~\cite{ouyang2022instructgpt,zhang2023reasoning}. They capture what experts say and do, but not how their $W^H_t$ and $G^H_t$ change through discrepancy-driven sensemaking. As a result, agents learn to imitate surface-level behaviour rather than participate in the \emph{chain of sensemaking}: joint discrepancy detection, causal explanation, goal refinement, and robust action.
\textbf{Future Directions.} CCS calls for richer \emph{sensemaking trajectories}: for \Stu{}, this means (context: repeated errors; anomaly: correct diagrams, wrong explanations; hypotheses: linguistic vs.\ procedural; disagreement: \Ag{} vs.\ \Tea{}; repair: vocabulary clarification; goal shift: deprioritise diagrams). Annotation schemes should distinguish \emph{epistemic actions} (\Ag{} surfacing a hypothesis, \Tea{} probing \Stu{}'s vocabulary) from \emph{instrumental actions} (executing a chosen plan)~\cite{lin2025intent}. Interactive fine-tuning protocols can log not only corrections but \emph{why} the expert thinks the assistant erred and how the expert's own model changed. Naturalistic logging (with governance) can capture genuine goal evolution.

Rather than generic multi-agent simulations, CCS points to \emph{constructivist collaborative playworlds} engineered as ``discrepancy engines'': environments that induce epistemic friction by giving agents partial, biased views of a shared process~\cite{baker2020emergent,zhou2024sotopia,zhou2024sotopiapi,park2023generative,qi2024civrealm}. Such playworlds annotate \emph{epistemic moves} (noticing mismatches, proposing causal links, renegotiating goals), turning sensemaking trajectories into supervision signals. In such playworlds (e.g., \Ag{} sees quiz logs; \Tea{} sees classroom behaviour; success requires negotiating a shared diagnosis of \Stu{}'s confusion), synthetic experts and assistants can be endowed with different $W$ and $G$ and must align them over time to succeed. Playworlds should be organised into \emph{curricula}: early levels exercise single-student, single-concept scenarios; later levels involve multi-student patterns, multi-week arcs, and stakeholder conflicts (\Tea{} wants depth; the principal wants test-score gains). Such curricula provide a concrete experimental path: they allow us to study when agents learn clarification, reframing, or goal renegotiation rather than one-shot prediction.

\textbf{Agenda 4: Architectures for Persistent, Structured Models.} LLM-based agents are typically stateless beyond short context windows. They lack persistent, structured world models $W^A_t$ that can be maintained across tasks, explicit representations of goals $G^A_t$ that can be revised, and memory systems that record when and why these structures changed. As a result, an agent may learn something important in one interaction and contradict it in the next, or treat transient objectives as if they were stable values.
\textbf{Future Directions.} CCS suggests architectural desiderata rather than a single blueprint. \emph{Neuro-symbolic causal twins} maintain explicit, editable models of the domain that both human and AI can inspect and revise (e.g., a graph where \Ag{} and \Tea{} inspect and edit nodes for \Stu{}, Newton's laws, and \Tea{}'s goals; when \Ag{} surfaces ``\Stu{}: confusion likely linguistic,'' \Tea{} can confirm or reject), serving as shared artefacts for sensemaking~\cite{vennix1996group,grieves2017digitaltwin}. In such architectures, LLMs serve as flexible ``epistemic encoders'' that translate language and observations into edits on an explicit causal and goal model, while a lightweight reasoner checks consistency, supports counterfactual prediction, and records provenance.

\emph{Episodic sensemaking memory} stores triplets such as (\Stu{} answered diagrams correctly; verbal explanations wrong; \Tea{} deprioritised diagram remediation), enabling \Ag{} to learn: ``correct-procedure-wrong-explanation triggers vocabulary investigation for \Tea{}.'' \emph{Teleological representations} such as reward machines~\cite{icarte2018rewardmachines} can encode the logical structure of goals; joint inference over these machines and causal graphs can link epistemic updates (editing $W$) to teleological updates (editing $G$). A lightweight \emph{theory-of-mind module} maintains hypotheses about $W^H_t$ and $G^H_t$; \Ag{} models \Tea{}'s preference for discovery, framing recommendations accordingly. In deployment, whether CCS behaviours actually manifest will depend on how these structures are threaded through context windows: episodic memories must survive across interactions and be retrieved at the right time to automatically shape future recommendations.

\textbf{Agenda 5: When to Disagree, When to Defer.} Even with appropriate objectives, data, and architectures, we lack principled policies for when CCS agents should agree, challenge, ask clarifying questions, or slow interaction for epistemic repair~\cite{Bansal2021UpdatesHarm,alonbarkat2023trust,li2025hacrl}. Current assistants are optimised for low-friction helpfulness: they answer quickly, avoid conflict, and rarely question the user's framing. Effective collaborators must sometimes do the opposite: pause, surface uncertainty, or propose goal revisions. At the same time, CCS introduces new risks: agents that infer and update goals endogenously may develop goal structures that drift away from human intent; agents trained to avoid sycophancy may become overconfident or manipulative.
\textbf{Future Directions.} Beyond \emph{what} to say, CCS raises questions about \emph{when} an agent should surface discrepancies and slow interaction for epistemic repair instead of answering fluently and moving on. Value-of-Information criteria~\cite{lu2024activereThinking} can estimate an \emph{expected benefit of repair}, trading off uncertainty reduction, outcome criticality, and friction cost (e.g., should \Ag{} interrupt to flag \Stu{}'s risk, or trust \Tea{}'s long-game pedagogy?). Mixed-initiative protocols can formalise turn-taking and control: when the assistant is allowed to override, when it must defer, and when it suggests after-action reviews. Training for ``intelligent disobedience'' can teach agents to contest risky decisions. If \Tea{} says ``just give \Stu{} the worksheet,'' \Ag{} might respond: ``The worksheet raises scores, but the confusion may resurface in momentum. Flag for review?''

CCS systems will need \emph{teleological constraints}: constitutional principles that bound goal formation and prevent agents from extrapolating goals in undesirable ways. Avoiding both sycophancy and ``sycophancy inversion'' requires adaptive personalisation based on expertise, context, and stakes. High-stakes sensemaking should be auditable via \emph{epistemic provenance} trails (e.g., ``\Tea{} rejected three drill recommendations; inferred: prefers discovery''; ``\Tea{} said use swimming for \Stu{}; inferred: contextualise to interests'')~\cite{hao2025beyondhitl}. These concerns connect CCS to broader debates on accountability and human-in-the-loop oversight in MAS.

%% file: sections/conclusion.tex
\section{Conclusion}

We have argued that making LLM-based agents into genuine teammates in MAS for \emph{decision support} requires shifting from behavioural alignment to \emph{collaborative causal sensemaking}: the joint construction, critique, and revision of shared world and goal models that underpin decisions. Rather than treating collaboration as an interface layer, CCS treats the human's evolving mental models and objectives as part of the decision state that agents must track, stress-test, and help refine. We sketched an agent-theoretic view in which epistemic and teleological alignment appear alongside task reward, and outlined research challenges in formalisation, measurement, playworld design, architectures, and interaction policies. The central hypothesis is that such alignment can reduce verification burden while enabling calibrated reliance and productive disagreement, with near-term footholds in CCS playworlds, causal-twin prototypes, and shadow-mode deployment where agents must demonstrate these behaviours under naturalistic conditions. Where instruction tuning builds tools that obey, CCS aims to build teammates that participate in the reasoning behind choices and \emph{think with} their human partners.

%% file: references.bib
@inproceedings{Bansal2021UpdatesHarm,
author = {Bansal, Gagan and Wu, Tongshuang and Zhou, Joyce and Fok, Raymond and Nushi, Besmira and Kamar, Ece and Ribeiro, Marco Tulio and Weld, Daniel},
title = {Does the Whole Exceed its Parts? The Effect of AI Explanations on Complementary Team Performance},
year = {2021},
isbn = {9781450380966},
publisher = {Association for Computing Machinery},
address = {New York, NY, USA},
url = {https://doi.org/10.1145/3411764.3445717},
doi = {10.1145/3411764.3445717},
booktitle = {Proceedings of the 2021 CHI Conference on Human Factors in Computing Systems},
articleno = {81},
numpages = {16},
location = {Yokohama, Japan},
series = {CHI '21}
}

@article{aha2018goalreasoning, 
title={Goal Reasoning: Foundations, Emerging Applications, and Prospects}, 
volume={39}, 
url={https://ojs.aaai.org/aimagazine/index.php/aimagazine/article/view/2800}, 
DOI={10.1609/aimag.v39i2.2800}, 
abstractNote={&lt;p class=&quot;Text&quot;&gt;Goal reasoning (GR) has a bright future as a foundation for the research and development of intelligent agents. GR is the study of agents that can deliberate on and self-select their goals/objectives, which is a desirable capability for some applications of deliberative autonomy. While studied in diverse AI sub-communities for multiple applications, our group has focused on how GR can play a key role for controlling autonomous systems. Thus, its importance is rapidly growing and it merits increased attention, particularly from the perspective of research on AI safety. In this article, I introduce GR, briefly relate it to other AI topics, summarize some of our group’s work on GR foundations and emerging applications, and describe some current and future research directions.&lt;/p&gt;}, 
number={2}, 
journal={AI Magazine}, 
author={Aha, David W.}, 
year={2018}, 
month={Jul.}, 
pages={3-24} }

@article{alonbarkat2023trust,
    author = {Alon-Barkat, Saar and Busuioc, Madalina},
    title = {Human–AI Interactions in Public Sector Decision Making: “Automation Bias” and “Selective Adherence” to Algorithmic Advice},
    journal = {Journal of Public Administration Research and Theory},
    volume = {33},
    number = {1},
    pages = {153-169},
    year = {2022},
    month = {02},
    issn = {1053-1858},
    doi = {10.1093/jopart/muac007},
    url = {https://doi.org/10.1093/jopart/muac007},
    eprint = {https://academic.oup.com/jpart/article-pdf/33/1/153/48511387/muac007.pdf},
}

@misc{awesome_llm_posttraining,
      title={A Survey on Post-training of Large Language Models}, 
      author={Guiyao Tie and Zeli Zhao and Dingjie Song and Fuyang Wei and Rong Zhou and Yurou Dai and Wen Yin and Zhejian Yang and Jiangyue Yan and Yao Su and Zhenhan Dai and Yifeng Xie and Yihan Cao and Lichao Sun and Pan Zhou and Lifang He and Hechang Chen and Yu Zhang and Qingsong Wen and Tianming Liu and Neil Zhenqiang Gong and Jiliang Tang and Caiming Xiong and Heng Ji and Philip S. Yu and Jianfeng Gao},
      year={2025},
      eprint={2503.06072},
      archivePrefix={arXiv},
      primaryClass={cs.CL},
      url={https://arxiv.org/abs/2503.06072}, 
}

@inproceedings{baker2020emergent,
title={Emergent Tool Use From Multi-Agent Autocurricula},
author={Bowen Baker and Ingmar Kanitscheider and Todor Markov and Yi Wu and Glenn Powell and Bob McGrew and Igor Mordatch},
booktitle={International Conference on Learning Representations},
year={2020},
url={https://openreview.net/forum?id=SkxpxJBKwS}
}

@article{bonawitz2011double,
title = {The double-edged sword of pedagogy: Instruction limits spontaneous exploration and discovery},
journal = {Cognition},
volume = {120},
number = {3},
pages = {322-330},
year = {2011},
note = {Probabilistic models of cognitive development},
issn = {0010-0277},
doi = {https://doi.org/10.1016/j.cognition.2010.10.001},
url = {https://www.sciencedirect.com/science/article/pii/S0010027710002258},
author = {Elizabeth Bonawitz and Patrick Shafto and Hyowon Gweon and Noah D. Goodman and Elizabeth Spelke and Laura Schulz}
}

@article{bucinca2024towards,
  author       = {Zana Bu{\c{c}}inca and
                  Siddharth Swaroop and
                  Amanda E. Paluch and
                  Susan A. Murphy and
                  Krzysztof Z. Gajos},
  title        = {Towards Optimizing Human-Centric Objectives in AI-Assisted Decision-Making
                  With Offline Reinforcement Learning},
  journal      = {CoRR},
  volume       = {abs/2403.05911},
  year         = {2024},
  url          = {https://doi.org/10.48550/arXiv.2403.05911},
  doi          = {10.48550/ARXIV.2403.05911},
  eprinttype    = {arXiv},
  eprint       = {2403.05911},
  bibsource    = {dblp computer science bibliography, https://dblp.org}
}

@incollection{cannonbowers1993shared,
  title={Shared mental models in expert team decision making},
  author={Janis A. Cannon-Bowers and Eduardo Salas and Sharolyn A. Converse},
  booktitle={Individual and Group Decision Making: Current Issues},
  editor={Castellan, N. John},
  year={1993},
  pages={221--246},
  publisher={Lawrence Erlbaum Associates},
  address={Hillsdale, NJ},
  url={https://api.semanticscholar.org/CorpusID:140519971}
}

@article{casper2023openproblems,
title={Open Problems and Fundamental Limitations of Reinforcement Learning from Human Feedback},
author={Stephen Casper and Xander Davies and Claudia Shi and Thomas Krendl Gilbert and J{\'e}r{\'e}my Scheurer and Javier Rando and Rachel Freedman and Tomek Korbak and David Lindner and Pedro Freire and Tony Tong Wang and Samuel Marks and Charbel-Raphael Segerie and Micah Carroll and Andi Peng and Phillip J.K. Christoffersen and Mehul Damani and Stewart Slocum and Usman Anwar and Anand Siththaranjan and Max Nadeau and Eric J Michaud and Jacob Pfau and Dmitrii Krasheninnikov and Xin Chen and Lauro Langosco and Peter Hase and Erdem Biyik and Anca Dragan and David Krueger and Dorsa Sadigh and Dylan Hadfield-Menell},
journal={Transactions on Machine Learning Research},
issn={2835-8856},
year={2023},
url={https://openreview.net/forum?id=bx24KpJ4Eb},
note={Survey Certification, Featured Certification}
}

@inproceedings{clark2025epistemic,
title={Epistemic Alignment: A Mediating Framework for User-{LLM} Knowledge Delivery},
author={Nicholas Clark and Hua Shen and Bill Howe and Tanu Mitra},
booktitle={Second Conference on Language Modeling},
year={2025},
url={https://openreview.net/forum?id=Orvjm9UqH2}
}

@inproceedings{cpo,
author = {Xu, Haoran and Sharaf, Amr and Chen, Yunmo and Tan, Weiting and Shen, Lingfeng and Van Durme, Benjamin and Murray, Kenton and Kim, Young Jin},
title = {Contrastive preference optimization: pushing the boundaries of LLM performance in machine translation},
year = {2024},
publisher = {JMLR.org},
booktitle = {Proceedings of the 41st International Conference on Machine Learning},
articleno = {2275},
numpages = {21},
location = {Vienna, Austria},
series = {ICML'24}
}

@book{craik1943nature,
  title={The Nature of Explanation},
  author={Kenneth Craik},
  year={1943},
  publisher={Cambridge University Press},
  address={Cambridge},
  url={https://api.semanticscholar.org/CorpusID:41364251}
}

@inproceedings{dvo,
author = {Ethayarajh, Kawin and Xu, Winnie and Muennighoff, Niklas and Jurafsky, Dan and Kiela, Douwe},
title = {Model alignment as prospect theoretic optimization},
year = {2024},
publisher = {JMLR.org},
booktitle = {Proceedings of the 41st International Conference on Machine Learning},
articleno = {504},
numpages = {18},
location = {Vienna, Austria},
series = {ICML'24}
}

@misc{fragiadakis2025evaluating,
      title={Evaluating Human-AI Collaboration: A Review and Methodological Framework}, 
      author={George Fragiadakis and Christos Diou and George Kousiouris and Mara Nikolaidou},
      year={2025},
      eprint={2407.19098},
      archivePrefix={arXiv},
      primaryClass={cs.HC},
      url={https://arxiv.org/abs/2407.19098}, 
}

@article{friston2024mentalmodels,
title = {Construction and use of mental models: Organizing principles for the science of brain and mind},
journal = {Neuropsychologia},
volume = {207},
pages = {109062},
year = {2025},
issn = {0028-3932},
doi = {https://doi.org/10.1016/j.neuropsychologia.2024.109062},
url = {https://www.sciencedirect.com/science/article/pii/S002839322400277X},
author = {John Duncan}
}

@article{goddard2012automation,
author = {Goddard, Kate and Roudsari, Abdul and Wyatt, Jeremy},
year = {2011},
month = {06},
pages = {121-7},
title = {Automation bias: A systematic review of frequency, effect mediators, and mitigators},
volume = {19},
journal = {Journal of the American Medical Informatics Association : JAMIA},
doi = {10.1136/amiajnl-2011-000089}
}

@ARTICLE{gomez2025taxonomy,
AUTHOR={Gomez, Catalina  and Cho, Sue Min  and Ke, Shichang  and Huang, Chien-Ming  and Unberath, Mathias },
TITLE={Human-AI collaboration is not very collaborative yet: a taxonomy of interaction patterns in AI-assisted decision making from a systematic review},
JOURNAL={Frontiers in Computer Science},    
VOLUME={6},
YEAR={2025},
URL={https://www.frontiersin.org/journals/computer-science/articles/10.3389/fcomp.2024.1521066},
DOI={10.3389/fcomp.2024.1521066},
ISSN={2624-9898}}

@article{gopnik2004causal,
	author = {Alison Gopnik and Clark Glymour and David M. Sobel and Laura E. Schulz and Tamar Kushnir and David Danks},
	doi = {10.1037/0033-295x.111.1.3},
	journal = {Psychological Review},
	number = {1},
	pages = {3--32},
	title = {A Theory of Causal Learning in Children: Causal Maps and Bayes Nets},
	volume = {111},
	year = {2004}
}

@inbook{grieves2017digitaltwin,
author = {Grieves, Michael and Vickers, John},
title = {Digital Twin: Mitigating Unpredictable, Undesirable Emergent Behavior in Complex Systems},
booktitle = {Transdisciplinary Perspectives on Complex Systems},
year = {2017},
month = {08},
pages = {85-113},
publisher = {Springer},
address = {Cham},
isbn = {978-3-319-38754-3},
doi = {10.1007/978-3-319-38756-7_4}
}

@inproceedings{hadfieldmenell2016cirl,
author = {Hadfield-Menell, Dylan and Dragan, Anca and Abbeel, Pieter and Russell, Stuart},
title = {Cooperative inverse reinforcement learning},
year = {2016},
isbn = {9781510838819},
publisher = {Curran Associates Inc.},
address = {Red Hook, NY, USA},
booktitle = {Proceedings of the 30th International Conference on Neural Information Processing Systems},
pages = {3916–3924},
numpages = {9},
location = {Barcelona, Spain},
series = {NIPS'16}
}

@inproceedings{hafner2020dreamer,
title={Dream to Control: Learning Behaviors by Latent Imagination},
author={Danijar Hafner and Timothy Lillicrap and Jimmy Ba and Mohammad Norouzi},
booktitle={International Conference on Learning Representations},
year={2020},
url={https://openreview.net/forum?id=S1lOTC4tDS}
}

@inproceedings{hao2023llmreasoners,
    title = "Reasoning with Language Model is Planning with World Model",
    author = "Hao, Shibo  and
      Gu, Yi  and
      Ma, Haodi  and
      Hong, Joshua  and
      Wang, Zhen  and
      Wang, Daisy  and
      Hu, Zhiting",
    editor = "Bouamor, Houda  and
      Pino, Juan  and
      Bali, Kalika",
    booktitle = "Proceedings of the 2023 Conference on Empirical Methods in Natural Language Processing",
    month = dec,
    year = "2023",
    address = "Singapore",
    publisher = "Association for Computational Linguistics",
    url = "https://aclanthology.org/2023.emnlp-main.507/",
    doi = "10.18653/v1/2023.emnlp-main.507",
    pages = "8154--8173"
}

@article{hao2025beyondhitl,
title = {Beyond human-in-the-loop: Sensemaking between artificial intelligence and human intelligence collaboration},
journal = {Sustainable Futures},
volume = {10},
pages = {101152},
year = {2025},
issn = {2666-1888},
doi = {https://doi.org/10.1016/j.sftr.2025.101152},
url = {https://www.sciencedirect.com/science/article/pii/S2666188825007166},
author = {Xinyue Hao and Emrah Demir and Daniel Eyers}
}

@inproceedings{
havrilla2024rlvr,
title={Teaching Large Language Models to Reason with Reinforcement Learning},
author={Alexander Havrilla and Yuqing Du and Sharath Chandra Raparthy and Christoforos Nalmpantis and Jane Dwivedi-Yu and Eric Hambro and Sainbayar Sukhbaatar and Roberta Raileanu},
booktitle={AI for Math Workshop @ ICML 2024},
year={2024},
url={https://openreview.net/forum?id=mjqoceuMnI}
}

@article{hemmer2024complementarity,
author = {Patrick Hemmer and Max Schemmer and Niklas Kühl and Michael Vössing and Gerhard Satzger},
title = {Complementarity in human-AI collaboration: concept, sources, and evidence},
journal = {European Journal of Information Systems},
volume = {34},
number = {6},
pages = {979--1002},
year = {2025},
publisher = {Taylor \& Francis},
doi = {10.1080/0960085X.2025.2475962},
URL = {https://doi.org/10.1080/0960085X.2025.2475962},
eprint = {https://doi.org/10.1080/0960085X.2025.2475962}}

@book{hovmand2014cbds,
author = {Hovmand, Peter S.},
year = {2013},
title = {Community Based System Dynamics},
isbn = {978-1-4614-8762-3},
publisher = {Springer},
doi = {10.1007/978-1-4614-8763-0}
}

@InProceedings{icarte2018rewardmachines,
  title = 	 {Using Reward Machines for High-Level Task Specification and Decomposition in Reinforcement Learning},
  author =       {Icarte, Rodrigo Toro and Klassen, Toryn and Valenzano, Richard and McIlraith, Sheila},
  booktitle = 	 {Proceedings of the 35th International Conference on Machine Learning},
  pages = 	 {2107--2116},
  year = 	 {2018},
  editor = 	 {Dy, Jennifer and Krause, Andreas},
  volume = 	 {80},
  series = 	 {Proceedings of Machine Learning Research},
  month = 	 {10--15 Jul},
  publisher =    {PMLR},
  url = 	 {https://proceedings.mlr.press/v80/icarte18a.html}
}

@book{johnsonlaird1983mental,
author = {Johnson-Laird, P. N.},
title = {Mental models: towards a cognitive science of language, inference, and consciousness},
year = {1986},
isbn = {0674568826},
publisher = {Harvard University Press},
address = {USA}
}

@article{kaelbling1998planning,
title = {Planning and acting in partially observable stochastic domains},
journal = {Artificial Intelligence},
volume = {101},
number = {1},
pages = {99-134},
year = {1998},
issn = {0004-3702},
doi = {https://doi.org/10.1016/S0004-3702(98)00023-X},
url = {https://www.sciencedirect.com/science/article/pii/S000437029800023X},
author = {Leslie Pack Kaelbling and Michael L. Littman and Anthony R. Cassandra}
}

@book{klein1998sources,
 ISBN = {9780262534291},
 URL = {http://www.jstor.org/stable/j.ctt1v2xt08},
 author = {Gary Klein},
 edition = {20},
 publisher = {The MIT Press},
 title = {Sources of Power: How People Make Decisions},
 urldate = {2026-01-13},
 year = {1998}
}

@ARTICLE{klein2006dataframe,
  author={Klein, G. and Moon, B. and Hoffman, R.R.},
  journal={IEEE Intelligent Systems}, 
  title={Making Sense of Sensemaking 2: A Macrocognitive Model}, 
  year={2006},
  volume={21},
  number={5},
  pages={88-92},
  doi={10.1109/MIS.2006.100}}

@inproceedings{lai2023designspace,
author = {Lai, Vivian and Chen, Chacha and Smith-Renner, Alison and Liao, Q. Vera and Tan, Chenhao},
title = {Towards a Science of Human-AI Decision Making: An Overview of Design Space in Empirical Human-Subject Studies},
year = {2023},
isbn = {9798400701924},
publisher = {Association for Computing Machinery},
address = {New York, NY, USA},
url = {https://doi.org/10.1145/3593013.3594087},
doi = {10.1145/3593013.3594087},
booktitle = {Proceedings of the 2023 ACM Conference on Fairness, Accountability, and Transparency},
pages = {1369–1385},
numpages = {17},
location = {Chicago, IL, USA},
series = {FAccT '23}
}

@article{li2025hacrl,
	author = {Li, Wei and Liu, Hongming and Huang, Kaizhu and Hussain, Amir},
	date = {2025/09/16},
	doi = {10.1007/s12559-025-10500-7},
	id = {Li2025},
	isbn = {1866-9964},
	journal = {Cognitive Computation},
	number = {5},
	pages = {146},
	title = {Reinforcement Learning for Human-AI Collaboration: Challenges, Mechanisms, and Methods},
	url = {https://doi.org/10.1007/s12559-025-10500-7},
	volume = {17},
	year = {2025}}

@inproceedings{
lin2025intent,
title={Reinforcement Learning for Human-{AI} Collaboration via Probabilistic Intent Inference},
author={Yuxin Lin and Seyede Fatemeh Ghoreishi and Tian Lan and Mahdi Imani},
booktitle={Reinforcement Learning Conference},
year={2025},
url={https://openreview.net/forum?id=u5bi4lzEYx}
}

@article{lu2024activereThinking,
title = "1+1>2? Information, Humans, and Machines",
author = "Tian Lu and Yingjie Zhang",
note = "Publisher Copyright: {\textcopyright} 2024 INFORMS.",
year = "2025",
month = mar,
doi = "10.1287/isre.2023.0305",
language = "English (US)",
volume = "36",
pages = "394--418",
journal = "Information Systems Research",
issn = "1047-7047",
publisher = "INFORMS Inst.for Operations Res.and the Management Sciences",
number = "1",
}

@inproceedings{luong2024reft,
    title = "{R}e{FT}: Reasoning with Reinforced Fine-Tuning",
    author = "Trung, Luong  and
      Zhang, Xinbo  and
      Jie, Zhanming  and
      Sun, Peng  and
      Jin, Xiaoran  and
      Li, Hang",
    editor = "Ku, Lun-Wei  and
      Martins, Andre  and
      Srikumar, Vivek",
    booktitle = "Proceedings of the 62nd Annual Meeting of the Association for Computational Linguistics (Volume 1: Long Papers)",
    month = aug,
    year = "2024",
    address = "Bangkok, Thailand",
    publisher = "Association for Computational Linguistics",
    url = "https://aclanthology.org/2024.acl-long.410/",
    doi = "10.18653/v1/2024.acl-long.410",
    pages = "7601--7614"
}

@article{lyell2017automation,
    author = {Lyell, David and Coiera, Enrico},
    title = {Automation bias and verification complexity: a systematic review},
    journal = {Journal of the American Medical Informatics Association},
    volume = {24},
    number = {2},
    pages = {423-431},
    year = {2016},
    month = {08},
    issn = {1067-5027},
    doi = {10.1093/jamia/ocw105},
    url = {https://doi.org/10.1093/jamia/ocw105},
    eprint = {https://academic.oup.com/jamia/article-pdf/24/2/423/34148461/ocw105.pdf},
}

@inproceedings{marivate2014quantifying,
  title={Quantifying Uncertainty in Batch Personalized Sequential Decision Making},
  author={Vukosi Marivate and Jessica J. Chemali and Emma Brunskill and Michael L. Littman},
  booktitle={AAAI Workshop: Modern Artificial Intelligence for Health Analytics},
  year={2014},
  url={https://api.semanticscholar.org/CorpusID:17589600}
}

@article{mathieu2000influence,
  title={The influence of shared mental models on team process and performance.},
  author={John E. Mathieu and Tonia S. Heffner and Gerald F Goodwin and Eduardo Salas and Janis A. Cannon-Bowers},
  journal={The Journal of applied psychology},
  year={2000},
  volume={85 2},
  pages={273-83},
  url={https://api.semanticscholar.org/CorpusID:10070771}
}

@article{mohammed2010teammental,
author = {Susan Mohammed and Lori Ferzandi and Katherine Hamilton},
title ={Metaphor No More: A 15-Year Review of the Team Mental Model Construct},
journal = {Journal of Management},
volume = {36},
number = {4},
pages = {876-910},
year = {2010},
doi = {10.1177/0149206309356804},
URL = {https://doi.org/10.1177/0149206309356804},
eprint = {https://doi.org/10.1177/0149206309356804}
}

@book{oliehoek2016decpomdp,
author = {Oliehoek, Frans A. and Amato, Christopher},
title = {A Concise Introduction to Decentralized POMDPs},
year = {2016},
isbn = {3319289276},
publisher = {Springer Publishing Company, Incorporated},
edition = {1st}
}

@inproceedings{ouyang2022instructgpt,
 author = {Ouyang, Long and Wu, Jeffrey and Jiang, Xu and Almeida, Diogo and Wainwright, Carroll and Mishkin, Pamela and Zhang, Chong and Agarwal, Sandhini and Slama, Katarina and Ray, Alex and Schulman, John and Hilton, Jacob and Kelton, Fraser and Miller, Luke and Simens, Maddie and Askell, Amanda and Welinder, Peter and Christiano, Paul F and Leike, Jan and Lowe, Ryan},
 booktitle = {Advances in Neural Information Processing Systems},
 editor = {S. Koyejo and S. Mohamed and A. Agarwal and D. Belgrave and K. Cho and A. Oh},
 pages = {27730--27744},
 publisher = {Curran Associates, Inc.},
 title = {Training language models to follow instructions with human feedback},
 url = {https://proceedings.neurips.cc/paper_files/paper/2022/file/b1efde53be364a73914f58805a001731-Paper-Conference.pdf},
 volume = {35},
 year = {2022}
}

@inproceedings{park2023generative,
author = {Park, Joon Sung and O'Brien, Joseph and Cai, Carrie Jun and Morris, Meredith Ringel and Liang, Percy and Bernstein, Michael S.},
title = {Generative Agents: Interactive Simulacra of Human Behavior},
year = {2023},
isbn = {9798400701320},
publisher = {Association for Computing Machinery},
address = {New York, NY, USA},
url = {https://doi.org/10.1145/3586183.3606763},
doi = {10.1145/3586183.3606763},
booktitle = {Proceedings of the 36th Annual ACM Symposium on User Interface Software and Technology},
articleno = {2},
numpages = {22},
location = {San Francisco, CA, USA},
series = {UIST '23}
}

@inproceedings{perez-etal-2023-discovering,
    title = "Discovering Language Model Behaviors with Model-Written Evaluations",
    author = "Perez, Ethan  and
      Ringer, Sam  and
      Lukosiute, Kamile  and
      Nguyen, Karina  and
      Chen, Edwin  and
      Heiner, Scott  and
      Pettit, Craig  and
      Olsson, Catherine  and
      Kundu, Sandipan  and
      Kadavath, Saurav  and
      Jones, Andy  and
      Chen, Anna  and
      Mann, Benjamin  and
      Israel, Brian  and
      Seethor, Bryan  and
      McKinnon, Cameron  and
      Olah, Christopher  and
      Yan, Da  and
      Amodei, Daniela  and
      Amodei, Dario  and
      Drain, Dawn  and
      Li, Dustin  and
      Tran-Johnson, Eli  and
      Khundadze, Guro  and
      Kernion, Jackson  and
      Landis, James  and
      Kerr, Jamie  and
      Mueller, Jared  and
      Hyun, Jeeyoon  and
      Landau, Joshua  and
      Ndousse, Kamal  and
      Goldberg, Landon  and
      Lovitt, Liane  and
      Lucas, Martin  and
      Sellitto, Michael  and
      Zhang, Miranda  and
      Kingsland, Neerav  and
      Elhage, Nelson  and
      Joseph, Nicholas  and
      Mercado, Noemi  and
      DasSarma, Nova  and
      Rausch, Oliver  and
      Larson, Robin  and
      McCandlish, Sam  and
      Johnston, Scott  and
      Kravec, Shauna  and
      El Showk, Sheer  and
      Lanham, Tamera  and
      Telleen-Lawton, Timothy  and
      Brown, Tom  and
      Henighan, Tom  and
      Hume, Tristan  and
      Bai, Yuntao  and
      Hatfield-Dodds, Zac  and
      Clark, Jack  and
      Bowman, Samuel R.  and
      Askell, Amanda  and
      Grosse, Roger  and
      Hernandez, Danny  and
      Ganguli, Deep  and
      Hubinger, Evan  and
      Schiefer, Nicholas  and
      Kaplan, Jared",
    editor = "Rogers, Anna  and
      Boyd-Graber, Jordan  and
      Okazaki, Naoaki",
    booktitle = "Findings of the Association for Computational Linguistics: ACL 2023",
    month = jul,
    year = "2023",
    address = "Toronto, Canada",
    publisher = "Association for Computational Linguistics",
    url = "https://aclanthology.org/2023.findings-acl.847/",
    doi = "10.18653/v1/2023.findings-acl.847",
    pages = "13387--13434"
}

@inproceedings{qi2024civrealm,
title={CivRealm: A Learning and Reasoning Odyssey in Civilization for Decision-Making Agents},
author={Siyuan Qi and Shuo Chen and Yexin Li and Xiangyu Kong and Junqi Wang and Bangcheng Yang and Pring Wong and Yifan Zhong and Xiaoyuan Zhang and Zhaowei Zhang and Nian Liu and Wei Wang and Yaodong Yang and Song-Chun Zhu},
booktitle={The Twelfth International Conference on Learning Representations},
year={2024},
url={https://openreview.net/forum?id=UBVNwD3hPN}
}

@inproceedings{rafailov2023dpo,
 author = {Rafailov, Rafael and Sharma, Archit and Mitchell, Eric and Manning, Christopher D and Ermon, Stefano and Finn, Chelsea},
 booktitle = {Advances in Neural Information Processing Systems},
 editor = {A. Oh and T. Naumann and A. Globerson and K. Saenko and M. Hardt and S. Levine},
 pages = {53728--53741},
 publisher = {Curran Associates, Inc.},
 title = {Direct Preference Optimization: Your Language Model is Secretly a Reward Model},
 url = {https://proceedings.neurips.cc/paper_files/paper/2023/file/a85b405ed65c6477a4fe8302b5e06ce7-Paper-Conference.pdf},
 volume = {36},
 year = {2023}
}

@article{rastogi2022taxonomy, 
title={A Taxonomy of Human and ML Strengths in Decision-Making to Investigate Human-ML Complementarity}, 
volume={11}, 
url={https://ojs.aaai.org/index.php/HCOMP/article/view/27554}, 
DOI={10.1609/hcomp.v11i1.27554}, 
number={1}, 
journal={Proceedings of the AAAI Conference on Human Computation and Crowdsourcing}, 
author={Rastogi, Charvi and Liu, Leqi and Holstein, Kenneth and Heidari, Hoda}, 
year={2023}, month={Nov.}, 
pages={127-139} }

@article{roijers2013survey,
author = {Roijers, Diederik M. and Vamplew, Peter and Whiteson, Shimon and Dazeley, Richard},
title = {A survey of multi-objective sequential decision-making},
year = {2013},
issue_date = {October 2013},
publisher = {AI Access Foundation},
address = {El Segundo, CA, USA},
volume = {48},
number = {1},
issn = {1076-9757},
journal = {J. Artif. Int. Res.},
month = oct,
pages = {67–113},
numpages = {47}
}

@inproceedings{sharma2024towards,
title={Towards Understanding Sycophancy in Language Models},
author={Mrinank Sharma and Meg Tong and Tomasz Korbak and David Duvenaud and Amanda Askell and Samuel R. Bowman and Esin DURMUS and Zac Hatfield-Dodds and Scott R Johnston and Shauna M Kravec and Timothy Maxwell and Sam McCandlish and Kamal Ndousse and Oliver Rausch and Nicholas Schiefer and Da Yan and Miranda Zhang and Ethan Perez},
booktitle={The Twelfth International Conference on Learning Representations},
year={2024},
url={https://openreview.net/forum?id=tvhaxkMKAn}
}

@article{steyvers2022bayesian,
author = {Mark Steyvers  and Heliodoro Tejeda  and Gavin Kerrigan  and Padhraic Smyth },
title = {Bayesian modeling of human–AI complementarity},
journal = {Proceedings of the National Academy of Sciences},
volume = {119},
number = {11},
pages = {e2111547119},
year = {2022},
doi = {10.1073/pnas.2111547119},
URL = {https://www.pnas.org/doi/abs/10.1073/pnas.2111547119},
eprint = {https://www.pnas.org/doi/pdf/10.1073/pnas.2111547119}}

@book{sutton2018reinforcement,
  author    = {Sutton, Richard S. and Barto, Andrew G.},
  title     = {Reinforcement Learning: An Introduction},
  edition   = {2nd},
  publisher = {{MIT} Press},
  address   = {Cambridge, MA},
  year      = {2018},
  isbn      = {9780262039246},
  url       = {http://incompleteideas.net/book/the-book-2nd.html}
}

@book{vennix1996group,
  title        = {Group Model Building: Facilitating Team Learning Using System Dynamics},
  author       = {Vennix, Jac A. M.},
  year         = {1996},
  publisher    = {Wiley},
  address      = {Chichester, UK}
}

@INPROCEEDINGS{wang2019bayesian,
  author={Wang, Xin and Kadioglu, Serdar},
  booktitle={2019 IEEE 31st International Conference on Tools with Artificial Intelligence (ICTAI)}, 
  title={Bayesian Deep Learning Based Exploration-Exploitation for Personalized Recommendations}, 
  year={2019},
  volume={},
  number={},
  pages={1715-1719},
  doi={10.1109/ICTAI.2019.00253}}

@inproceedings{wei2022chainofthought,
author = {Wei, Jason and Wang, Xuezhi and Schuurmans, Dale and Bosma, Maarten and Ichter, Brian and Xia, Fei and Chi, Ed H. and Le, Quoc V. and Zhou, Denny},
title = {Chain-of-thought prompting elicits reasoning in large language models},
year = {2022},
isbn = {9781713871088},
publisher = {Curran Associates Inc.},
address = {Red Hook, NY, USA},
booktitle = {Proceedings of the 36th International Conference on Neural Information Processing Systems},
articleno = {1800},
numpages = {14},
location = {New Orleans, LA, USA},
series = {NIPS '22}
}

@article{wei2022emergent,
title={Emergent Abilities of Large Language Models},
author={Jason Wei and Yi Tay and Rishi Bommasani and Colin Raffel and Barret Zoph and Sebastian Borgeaud and Dani Yogatama and Maarten Bosma and Denny Zhou and Donald Metzler and Ed H. Chi and Tatsunori Hashimoto and Oriol Vinyals and Percy Liang and Jeff Dean and William Fedus},
journal={Transactions on Machine Learning Research},
issn={2835-8856},
year={2022},
url={https://openreview.net/forum?id=yzkSU5zdwD},
note={Survey Certification}
}

@book{weick1995sensemaking,
  title        = {Sensemaking in Organizations},
  author       = {Weick, Karl E.},
  year         = {1995},
  publisher    = {SAGE},
  address      = {Thousand Oaks, CA}
}

@misc{zelikman2024quietstar,
      title={Quiet-STaR: Language Models Can Teach Themselves to Think Before Speaking}, 
      author={Eric Zelikman and Georges Harik and Yijia Shao and Varuna Jayasiri and Nick Haber and Noah D. Goodman},
      year={2024},
      eprint={2403.09629},
      archivePrefix={arXiv},
      primaryClass={cs.CL},
      url={https://arxiv.org/abs/2403.09629}, 
}

@inproceedings{zhang2020appropriatereliance,
author = {Zhang, Yunfeng and Liao, Q. Vera and Bellamy, Rachel K. E.},
title = {Effect of confidence and explanation on accuracy and trust calibration in AI-assisted decision making},
year = {2020},
isbn = {9781450369367},
publisher = {Association for Computing Machinery},
address = {New York, NY, USA},
url = {https://doi.org/10.1145/3351095.3372852},
doi = {10.1145/3351095.3372852},
booktitle = {Proceedings of the 2020 Conference on Fairness, Accountability, and Transparency},
pages = {295–305},
numpages = {11},
location = {Barcelona, Spain},
series = {FAT* '20}
}

@inproceedings{zhou2024sotopia,
title={{SOTOPIA}: Interactive Evaluation for Social Intelligence in Language Agents},
author={Xuhui Zhou and Hao Zhu and Leena Mathur and Ruohong Zhang and Haofei Yu and Zhengyang Qi and Louis-Philippe Morency and Yonatan Bisk and Daniel Fried and Graham Neubig and Maarten Sap},
booktitle={The Twelfth International Conference on Learning Representations},
year={2024},
url={https://openreview.net/forum?id=mM7VurbA4r}
}
